david@algolabs.com
sian@algolabs.com

# Learning From String Sequences

David Lindsay[1] and Siân Cox[2]

[1] Computer Learning Research Centre, Royal Holloway, University of London,
Egham, Surrey, TW20 OEX, UK,
`davidl@cs.rhul.ac.uk`
[2] School of Biological Sciences, Royal Holloway, University of London, Egham,
Surrey, TW20 OEX, UK, `s.s.e.cox.rhul.ac.uk`

**Abstract.** The Universal Similarity Metric (USM) has been demonstrated to give practically useful measures of "similarity" between sequence data. Here we have used the USM as an alternative distance metric in a $K$-Nearest Neighbours ($K$-NN) learner to allow effective pattern recognition of variable length sequence data. We compare this USM approach with the commonly used string-to-word vector approach. Our experiments have used two data sets of divergent domains: (1) spam e-mail filtering and (2) protein subcellular localisation. Our results with this data reveal that the USM based $K$-NN learner (1) gives predictions with higher classification accuracy than those output by techniques that use the string to word vector approach, and (2) can be used to generate reliable probability forecasts.

## 1 Introduction

The traditional approach to pattern recognition in the setting of supervised learning is dependent on creating a dataset which encapsulates the problem to be learned. The power of the eventual learning algorithm used to analyse the data is heavily limited by how and which features are used in the dataset. Many machine learning algorithms can be thought of as "number crunching" techniques that rely on the input data being represented as a fixed number of real valued features. For example, the $K$-Nearest Neighbours algorithm [1] (like many others) analyses given data by considering data examples as vectors in an multidimensional Euclidean space. For many applications modelling the problem in a numeric format is perfectly acceptable, however with more complex domains such as text and biological data the mixture of symbolic and real features is problematic. Because of this much research has focused on the design of alternative distance metrics (or kernels) that incorporate domain knowledge [2–5]. A commonly used approach is to break down data sequences into a dictionary of commonly occurring "words" [6]. For example, with sequences of English text these could be the actual words that comprise the sentences. The word frequency is then used as a numerical feature in constructing a new data set. This conversion into word frequencies often loses the effective context in which the words were originally used in the sequence .

Recent work by Li et al (2003) gave an elegant solution to computing an effective similarity measure between sequences. The "Universal Similarity Metric" (USM) is based on the non-computable notion of Kolmogorov Complexity. The USM measure is a provenly universal normalised information distance which allows comparison and recognition of all effective similarities between variable length sequences. A key advantage of the USM is that it analyses the complete sequence of words in their original ordered context.The practical application of the USM to problems in bioinformatics and linguistics has previously been described [7]. We have extended the use of the USM as an alternative distance metric for use in the $K$-NN algorithm.

To demonstrate the effectiveness of the USM based $K$-NN algorithm we have used two divergent and complex problems of sequence classification:

1. **Spam filtering** Electronic mail is a popular and powerful communications medium. However the increasing volume of unsolicited bulk e-mail (spam) has generated a need for reliable anti-spam filters [8].
2. **Protein localisation** The assignment of function to a protein is difficult when no clear homology of a protein with known functions exist. Knowing which subcellular location a protein resides may give important insights into its function [9].

With both problems we experimented using publicly available real life data sets. Both data sets had already been analysed using traditional machine learning approaches [8, 10, 9, 11]. Experiments with the USM $K$-NN algorithm show improved prediction accuracy over the word-to-vector approaches when tested on the spam data set and comparable results with the protein data set. We show that the USM $K$-NN algorithm can effectively analyse sequence data of variable lengths, where the raw sequence data can be analysed directly without any pre-formatting. In addition, we show that no modification to the algorithm is required to apply it to either data set tested, despite significant differences in problem domains. Finally, using empirical reliability curves and loss functions we show that the use of USM in a $K$-NN learner allows simple generation of probability forecasts which are more reliable than those of other learners such as Naive Bayes.

## 2 The Universal Similarity Metric (USM)

### 2.1 Theoretical definition

We used a mathematical definition of the USM given by [7] as follows. Given two sequences $x$ and $y$, we define the USM distance between the sequences $d(x,y)$ by

$$d(x,y) = \frac{\max\{K(x \mid y^*), K(y \mid x^*)\}}{\max\{K(x), K(y)\}} \quad (1)$$

The Kolmogorov complexity $K(x)$ of a binary sequence $x$ is defined as the shortest program needed to run on a Universal Turing Machine to recreate sequence $x$ on its output tape. Let $y^*$ denote the shortest program for $y$. The conditional Kolmogorov complexity $K(x \mid y^*)$ is defined as the length of the shortest Universal Turing Machine program that can reproduce sequence $x$ given sequence $y*$ on an auxiliary input tape.

The USM has been shown by Li et al (2003) to take values in $[0, 1]$ and be a valid distance metric, satisfying the triangular inequality (up to asymptotic vanishing additive error terms). These theoretical properties make the metric suitable for use in the $K$-Nearest Neighbours component of the USM $K$-NN algorithm.

## 2.2 Practical implementation

As with many concepts born from the Theory of Algorithmic randomness, the notion of Kolmogorov complexity is non-computable. We used compression algorithms to approximate the Kolmogorov complexity of a given sequence. For example, if a fixed compression program which compresses a sequence $x$ to another sequence $x'$, we have $K(x) \leq |x'| \leq |x|$. Therefore the length of the compressed version $|x'|$ of the sequence can be used as an approximation of $K(x)$.

To compute $K(x \mid y^*)$ we use a deep theorem in Kolmogorov Complexity called the Symmetry of Information that changes the problem to:

$$K(x \mid y^*) = K(x, y^*) - K(y^*) \qquad (2)$$

We replace $K(x, y^*)$ by $K(xy^*)$ which is accurate up to precision $K(|x|) \approx \log(|x|)$. The term $K(xy^*)$ is the Kolmogorov complexity of the two sequences $x$ and $y^*$ concatenated together.

We used simple Lempel Ziv compression algorithms provided with the Java programming language's Deflater/Inflater classes. This compression algorithm is a loss-less variant which means that the original string can be retrieved by running the algorithm in reverse on the compressed string. To compress each string it learns a dictionary as it compresses. With our experiments we used this same compression program on the email messages and the protein sequences. In general it is better practice to use an application specific compression program to compress the sequences as thoroughly as possible so that the approximation of the Kolmogorov complexity of the sequences is as close as possible. For example, programs like gzip could be used for English plain text and GenCompress for DNA sequences, as demonstrated in the experiments by Li et al (2003).

## 2.3 Alternative distance metric for $K$-NN learners

The main focus of this study will be to test the applicability of the USM as a distance measure for a $K$-NN learner. Many different distance metrics have been proposed for $K$-NN learners [2–5]. Of particular interest is the string-to-word vector approach, which converts strings into word frequencies. This approach easily allows many traditional learning algorithms to be used that are ideally suited to real valued data.

## 3 Experimental results

### 3.1 Description of the Ling-Spam data set

The Ling-Spam corpus is freely available from the web site

`http://www.iit.demokritos.gr/~ionandr`

The corpus contains 2893 example messages in total with 2 possible classes:

- 2412 legitimate emails
- 481 spam messages

Spam messages comprise 16.6% of the corpus, which reflect typical incoming spam rates [12]. Four different versions of the data set are available, each with different levels of data pre-processing. Level 1 data is the original email messages prior to preprocessing. Level 2 removes all "stop" words from messages. Stop words are defined to occur so frequently within messages that they fail to give any useful information about the type of email (e.g. words such as 'the', 'of' and 'but'). Level 3 applies a lemmatiser to the text. The lemmatiser converts all words of similar meaning into the same form (e.g. "earning" becomes "earn"). Level 4 removes all stop words and applies the lemmatiser to the text.

The email messages are of considerably different lengths. For level 1 data, the shortest message is 47 ASCII characters in length, the longest 28683, the average 3184 and the standard deviation 3549.

### 3.2 Spam offline accuracy results

To compare the accuracy of prediction of the USM $K$-NN algorithm against previous results as cited by Androutsopoulos *et al* (2000) we used their performance score systems defined as follows:

$$\text{Spam Recall} = \frac{n_{spam \to spam}}{N_{spam}} \qquad (3)$$

$$\text{Spam Precision} = \frac{n_{spam \to spam}}{n_{spam \to spam} + n_{legit \to spam}} \qquad (4)$$

The term $N_{spam}$ is the total number of email messages in the test set which have true label as *spam*. Intuitively the terms $n_{legit \to spam}$ are the number of test examples which have true label *legit* but are misclassified by the learning machine as *spam*. Similarly $n_{spam \to spam}$ represents the number of *spam* messages in the test set which are correctly classified as *spam*.

The three algorithms used by Androutsopoulos *et al* (2000) and their corresponding performance scores are shown in Table 1. All three algorithms used the occurrence of key words in each message as features to classify the email messages. These results were obtained using level 4 filtered data which allowed better statistics to be gathered from the data [8].

**Table 1.** Classification accuracies of the USM based $K$-NN learner and various other learning algorithms when tested on the Ling-Spam data set run with 10-fold cross validation.

| Algorithm | Spam Recall (%) | SpamPrec (%) |
|---|---|---|
| USM-1-NN | 92.5 | 99.1 |
| USM-10-NN | 95.01 | 98.7 |
| USM-20-NN | 95.43 | 99.14 |
| USM-30-NN | 94.8 | 98.49 |
| Naive Bayes [8] | 82.35 | 99.02 |
| TiMBL KNN (Trad 1NN) [8] | 85.27 | 95.92 |
| MS Outlook patterns [8] | 53.01 | 87.93 |

Using the USM based $K$-NN algorithm we performed 10-fold cross validation tests using the original folds that the Ling-Spam data were already divided into (Table 1). The USM $K$-NN algorithm Spam Recall and Precision performance scores were similar for all levels 1, 2, 3 and 4 of the data sets tested. The USM based $K$-NN algorithm gave highest predictive accuracy on the level 1 data (Table 1). This demonstrates that the USM works more effectively given the sequences of words in their original form; pre-processing of the data is not necessary for the USM based learner. The USM $K$-NN algorithms give significantly higher Spam Recall scores as compared with techniques previously described, e.g. USM-20-NN 92.9% as compared with TiMBL 85.27%. There is significant improvement of the USM $K$-NN learners over their non-USM counterparts, indicating that the USM distance is far better suited to the data than the Euclidean distance.

### 3.3 Protein results

We tested our USM $K$-NN and non-USM $K$-NN algorithms on the eukaryotic protein data set using a randomly split 10-Fold cross validation. The classification accuracies of these experiments are shown in Table 2 in addition to results previously described for other algorithms tested on the same data set [10, 11]. All previously described approaches use variants of the string to word vector approach also incorporating extra domain knowledge. For our standard $K$-NN learners we used the Euclidean distance between 20-dimensional vectors (counting the number of amino acids used in each protein).

Almost all USM $K$-NN learners improved prediction accuracy over the Kohonen SOM [10], Neural Network and Markov Model classifiers [11]. The results of SVM slightly improve on the USM-KNN result (overall accuracy 79.4% as compared with 74.0%); however the SVM is specifically designed for this application via the use of specialised kernels which incorporate domain knowledge.

**Table 2.** Classification accuracies of various learning algorithms on the eukaryotic protein localisation data set in the offline setting.

| Algorithm | Cyto (%) | Extra (%) | Mito (%) | Nucl (%) | Overall (%) |
|---|---|---|---|---|---|
| USM 1-NN | 76.6 | 73.8 | 50.5 | 84.8 | 76.5 |
| USM 10-NN | 83.9 | 79.1 | 53.6 | 76.4 | 75.9 |
| USM 20-NN | 83.6 | 79.1 | 53.6 | 76.5 | 75.8 |
| USM 30-NN | 85.9 | 88.0 | 53.9 | 70.7 | 74.0 |
| Naive Bayes | 76.0 | 73.8 | 49.8 | 82.7 | 75.3 |
| Kohonen SOM [10] | 72.1 | 71.4 | 43.3 | 77.4 | 70.6 |
| Neural Net [11] | 55.0 | 75.0 | 61.0 | 72.0 | 66.0 |
| Markov Model [11] | 78.1 | 62.2 | 69.2 | 74.1 | 73.0 |
| SVM [11] | 76.9 | 80.0 | 56.7 | 87.4 | 79.4 |

**Table 3.** Example output of predictions made by the USM 30-NN algorithm on the Eukaryotic protein localisation data set

| Seq No. | Protein Name | Real Class | Pred Class | Probability Forecasts $\hat{P}(y \mid x)$ | | | |
|---|---|---|---|---|---|---|---|
| | | | | (1) Cy | (2) Ex | (3) Mi | (4) Nu |
| 1 | EFGM_RAT | (3) Mi | (3) Mi | 2.423E-4 | 2.423E-4 | 0.709 | 0.290 |
| 2 | ARY2_HUMAN | (1) Cy | (1) Cy | 0.677 | 0.161 | 0.161 | 0.0 |
| 3 | PPT_BOVIN | (2) Ex | (2) Ex | 0.1 | 0.899 | 2.347E-4 | 2.347E-4 |
| 4 ◇ | RPC1_GIALA | (4) Nu | (1) Cy | 0.515 | 2.504E-4 | 2.504E-4 | 0.454 |
| 5 | ZEST_DROME | (4) Nu | (4) Nu | 0.066 | 2.347E-4 | 2.347E-4 | 0.933 |

### 3.4 Evaluation of probability forecasts

One advantage of using the USM in a $K$-NN learner is that conditional probability estimates can be obtained for each class label. For each test instance a set of $K$-closest training objects (in terms of USM distance) is found. The number of examples with each class label $K_y$ is used to obtain an estimate of the conditional probability $\hat{P}(y \mid x) = \frac{K_y}{K}$ of each possible label $y$ given the new test object $x$. This approach of estimating probabilities using a $K$-NN learner has proven convergence properties to the Bayes optimal distribution, however this phenomena can only be guaranteed asymptotically [13]. The results in Table 3 give the probability forecasts output by the USM-30-NN learner on examples selected from the Eukaryotic protein dataset. With correctly classified examples (sequence numbers 1, 2, 3 and 5) the probability forecasts are distinctly higher than all others on the correct classifications (e.g. sequence 2 has estimated probability = 0.677 for its real class Cytoplasm, and all other forecasts are less than 0.161).

The USM-30-NN algorithm incorrectly classifies sequence 4 as Cytoplasmic (marked ◇). Yet closer inspection of the corresponding forecasts for all other possible classes reveals that the next most likely class is Nuclear (the true class). In fact there is only a small difference between the true and predicted forecast values (0.515 Cytoplasmic and 0.454 Nuclear).

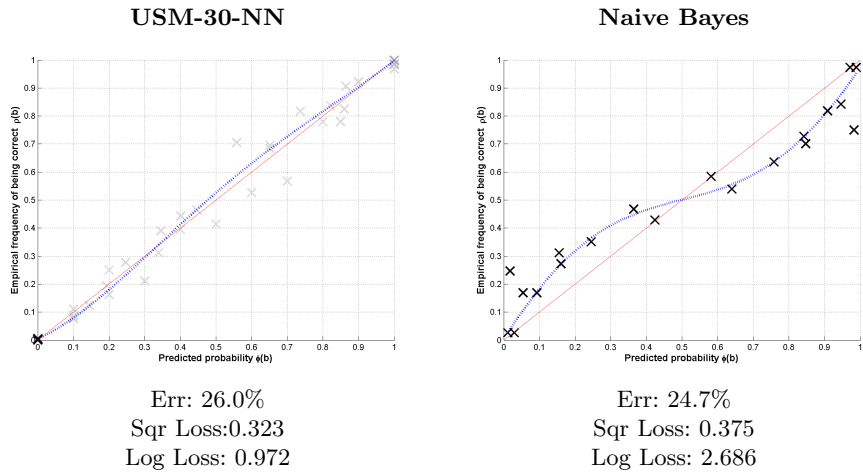

**Fig. 1.** Comparison of ERC plots of probability forecasts output by the USM-30-NN and Naive Bayes learners when tested on Eukaryotic protein dataset. The diagonal red line represents the *line of calibration*, where predicted probability (*horizontal*) equals observed fraction of correct predictions (*vertical*). Under- and over-estimation of probability forecasts are represented by the reliability curve (dotted blue line) deviating above and below the line of calibration respectively.

We deem probability forecasts as reliable if they do not lie. When a probability $\hat{p}$ is assigned to an event, there should be roughly $1-\hat{p}$ chance of the event not occurring [?]. Empirical reliability curves (ERC) [14] allow visualisation of the reliability of probability forecasts. ERC plots demonstrate those learners that display over- and under-estimation in their probability forecasts by way of the plot falling above or below the line of calibration. Figure 1 shows ERC plots of probability forecasts output by the USM-30-NN and Naive Bayes learners when tested on the Eukaryotic protein dataset. Both of these learners give comparable classification accuracy (Table 2), yet from the ERC plots it is obvious that Naive Bayes is most unreliable as it displays gross over- and under-estimation. On the other hand, the USM-30-NN learner produces very reliable probability forecasts as is indicated by the ERC plot aligning well with the ideal line of calibration. This point is also reiterated by the square and log loss scores for both learners (Figure 1), where less loss is incurred for the USM-30-NN. The unreliability of the Naive Bayes probability forecasts could be because the assumption of independence of attributes that this learner makes about the underlying probability distribution is violated [15]: the frequencies of amino acids are almost certainly correlated. In contrast, the USM-$K$-NN makes the weaker assumption that the data are i.i.d. and therefore produces more reliable probability forecasts [13].

## 4  Discussion

The USM-$K$-NN learner has been demonstrated to cope well with both data sets which consist of highly variable length sequence data. The USM is designed to be a normalised information distance; sequence length does not affect the measure. We suggest that the USM-$K$-NN algorithm gives high classification accuracy because it can interpret the data in its original format. Hence the USM-$K$-NN is able to exploit all the information in the dataset. It does not require the conversion of the data into numeric format. However, whilst the string-to-word vector approach enables quick processing of the data it often loses the order and context in which the words were originally used. In contrast, the USM utilises compression algorithms to generate a distance measure between strings, therefore the order in which the words occur is reflected in the distance metric. The email messages and protein sequences comprise similar symbols with very different meanings. Protein sequences use an 'alphabet' of 20 English letters, which are also present in the email messages. Despite these complexities, the USM-$K$-NN learner effectively realises similarity in both data without any changes to the algorithm. This means that user intervention to model domain specific knowledge is not required. As stated previously, possible improvements could be made by using a compression program more specific for protein sequences, however this is equivalent to modelling domain specific knowledge.

This study has shown that in addition to good classification accuracy, the USM-$K$-NN is able to produce reliable probability forecasts. The combination of good accuracy and reliability is an obvious practical advantage, since it allows the user to know when to trust the predictions output by the USM-$K$-NN learner. The reliable properties of the USM-$K$-NN learner can be attributed to the $K$-NN component: this learner has much theoretical and empirical evidence supporting its reliability [13]. The fact that the USM causes no detrimental factor to the reliability of the $K$-NN is a clear advantage. In principal the USM could be applied to many other learners, the most natural choice would be to introduce it into SVM as a kernel. However this would require further techniques to extract reliable probability forecasts [14].

The concept of utilising the concepts of Kolmogorov complexity as a distance metric for $K$-NN learners is not new; the KStar learner developed by Cleary et al (1995) utilises an entropic distance measure, estimating the transformations required to convert one instance to another. KStar offers a consistent approach to handling of symbolic attributes, real valued attributes and missing values. The problem with comparing such approaches with our USM-$K$-NN approach is that the KStar algorithm requires examples to be of a fixed vector size. To capitalise on this technique we would therefore have to break up the variable length strings into fixed length attribute vectors using a windowing approach [3]. For future work it would be interesting to compare the USM-$K$-NN and KStar algorithms.

We believe that this research has interesting parallels with common belief held in cognitive science that learning can be viewed as compression.

http://www.cognitionresearch.org.uk//cognition.html

The USM provides an interesting focal point for designing kernels and distance metrics. Current research is investigating the effects of using lossy compression instead of loss-less compression in USM implementations. We envisage that the USM technique will provide a new way of learning from multi-media data. Experiments are under way to see if learning from music and video files can be achieved using MP3 or DivX compression respectively[16].

## Acknowledgements

We thank Paul Vitanyi for explaining the concepts of the USM. We would like to acknowledge Roy Maunders, David Collins, Alex Gammerman and Volodya Vovk for their help in reviewing our work. This work was funded by an EPSRC Studentship Grant to DL.